\documentclass{article}
\usepackage{iclr2026_conference,times}
\usepackage[utf8]{inputenc}
\usepackage[T1]{fontenc}
\usepackage{amsmath,amssymb,amsfonts}
\usepackage{graphicx}
\usepackage{booktabs}
\usepackage{array}
\usepackage{hyperref}
\usepackage{url}
\usepackage{algorithm}
\usepackage{algorithmic}
\usepackage{multirow}
\usepackage[table]{xcolor}
\usepackage{subcaption}

\definecolor{vsigreen}{HTML}{2ecc71}
\definecolor{outcomeorange}{HTML}{f39c12}
\definecolor{baselinered}{HTML}{e74c3c}
\definecolor{vsirow}{HTML}{EAFAF1}    
\definecolor{peakcell}{HTML}{ABEBC6}  

\title{Reliable Self-Improvement Training by Verifying Reasoning, Not Just Answers}

\iclrfinalcopy

\author{
Xinyu Zhang \\
Anyscale \\
\texttt{xinyzng@gmail.com}
}

\newcommand{\vsi}{\textsc{VSI}}

\begin{document}

\maketitle

\begin{abstract}
Self-improvement training---where a model learns from solutions it generated itself---promises sustained capability gains, but it has a well-known failure mode: as a model trains on its own outputs round after round, small errors in its reasoning compound until accuracy stalls or degrades. We trace this \emph{drift} to \emph{what} gets kept. Most self-improvement methods keep a solution whenever its final answer is correct, which lets through ``lucky guesses''---answers that come out right despite flawed reasoning---and training on them teaches the flawed reasoning. We propose \textbf{Verified Self-Improvement (\vsi{})}, which keeps a self-generated solution only when its \emph{reasoning} holds up, not just its answer. Before a solution can enter the training set, \vsi{} recomputes every arithmetic step with a computer-algebra library (\texttt{sympy}), checks that intermediate values stay consistent from one step to the next, and enforces simple domain constraints (for example, counts should be non-negative whole numbers). We evaluate \vsi{} on GSM8K with Qwen3-4B-Thinking across 5 rounds of self-improvement training under five conditions: no verification, outcome (answer-only) verification, majority voting, full \vsi{} step-level checking, and \vsi{} with DPO. \vsi{} rejects about 34\% of the correct-answer solutions that outcome verification would keep---exactly the lucky guesses---and this cleaner training signal lets accuracy keep climbing across all five rounds (80.5\%$\to$91.0\%), whereas outcome verification plateaus and no verification collapses. Turning these checks into DPO preference pairs---preferring sound reasoning over lucky answers---further teaches the model to tell the two apart (reward accuracy 46\%$\to$63\%). \vsi{} is a simple, reproducible recipe for making self-improvement training reliable wherever an automatic check on the reasoning is available.
\end{abstract}

\section{Introduction}
\label{sec:intro}

Self-improvement training---where a model improves by learning from data it generates itself---is one of the most appealing ideas in modern machine learning. Recent work on self-training \citep{zelikman2022star, singh2024rest}, self-rewarding language models \citep{yuan2024selfrewarding}, and bootstrapped reasoning \citep{hosseini2024vstar} shows that a model can meaningfully improve from its own outputs through a simple loop: sample candidate solutions, keep the good ones, and finetune on them.

In practice, though, the loop is fragile. When a model is finetuned on its own outputs round after round, errors in intermediate reasoning compound from one round to the next---a problem we call \textbf{drift}. A model may reach a correct final answer through faulty reasoning (a ``lucky guess''); once that solution is kept and trained on, the model learns to reproduce the faulty reasoning, and over several rounds this builds into systematic degradation---the same dynamic behind model collapse when models are trained repeatedly on generated data \citep{shumailov2024model}.

Our central claim is simple: \emph{what you keep determines how long self-improvement keeps paying off}. Standard self-training keeps a solution based on its outcome alone---is the final answer correct?---which lets through solutions whose reasoning contains arithmetic slips, inconsistent intermediate values, or violated constraints that happen to cancel out. Finetuning on those solutions spreads the flawed reasoning.

We propose \textbf{Verified Self-Improvement (\vsi{})}: a self-improvement training loop that uses an automatic check on each solution's \emph{reasoning}---not just its answer---to decide what to finetune on. Concretely, \vsi{} is rejection-sampling finetuning with a stricter acceptance rule. Before a self-generated solution is kept, \vsi{} requires that:
\begin{enumerate}
    \item \textbf{The answer is correct}: the final answer matches the ground truth.
    \item \textbf{The arithmetic holds up}: every arithmetic step we can parse from the chain-of-thought is recomputed with \texttt{sympy} \citep{meurer2017sympy} and confirmed correct.
    \item \textbf{The reasoning is internally consistent}: intermediate values are tracked across steps, so we can catch a number the model invents that contradicts its own earlier work.
    \item \textbf{Basic constraints hold}: simple domain rules---non-negative counts, whole numbers for discrete quantities---are respected.
\end{enumerate}

By asking that \emph{every reasoning step}, not just the final answer, hold up, \vsi{} keeps lucky guesses out of the finetuning data, so the model learns from solutions that are right \emph{for the right reasons}. This is what keeps self-improvement training stable over more rounds.

\paragraph{Contributions.} (1) We introduce \vsi{}, a self-improvement training recipe that improves stability by filtering self-generated finetuning data on the quality of the \emph{reasoning}, not just the answer. (2) We show this acceptance rule is far more selective than outcome-only filtering ($\sim$52\% vs.\ $\sim$78\% of solutions kept), removing solutions that are right for the wrong reasons. (3) We show that the same checks yield DPO preference pairs that teach the model to prefer sound reasoning over lucky answers. (4) We provide a complete, reproducible pipeline for self-improvement training with automatic verification.

\section{Related Work}
\label{sec:related}

\paragraph{Self-training for reasoning.} STaR \citep{zelikman2022star} bootstraps reasoning by finetuning on self-generated rationales that reach correct answers. ReST-EM \citep{singh2024rest} casts this as an EM-style loop with repeated sampling. V-STaR \citep{hosseini2024vstar} trains verifiers alongside generators, and self-rewarding language models \citep{yuan2024selfrewarding} use the model itself as a judge. These methods accept solutions mainly by outcome correctness; \vsi{} adds an automatic check on the reasoning steps, improving the quality of the finetuning data.

\paragraph{Process supervision.} \citet{lightman2024lets} show that supervising each reasoning step beats supervising only the final answer when training math models, and \citet{uesato2022process} compare process and outcome feedback. These works rely on human-annotated step labels; \vsi{} obtains a step-level signal automatically by recomputing the math, at no annotation cost.

\paragraph{Tool-assisted and verified reasoning.} TORA \citep{gou2024tora} weaves tool use, including exact computation, into the reasoning process, and generative theorem provers \citep{polu2020generative} rely on formal verification. \vsi{} differs in \emph{where} the check is applied: it filters the finetuning data rather than acting at inference time, which makes it complementary to these approaches.

\paragraph{Model collapse.} \citet{shumailov2024model} show that repeatedly training on model-generated data progressively degrades quality, and \citet{huang2023large} show that LLMs struggle to self-correct without external feedback. \vsi{} supplies exactly that external signal---an automatic check on the reasoning---which keeps errors from propagating across rounds.

\section{Method: Verified Self-Improvement}
\label{sec:method}

\subsection{The Self-Improvement Training Loop}

\vsi{} runs as a self-improvement training loop (Figure~\ref{fig:pipeline}). Starting from a base model $M_0$, each round $i$ proceeds as:

\begin{enumerate}
    \item \textbf{Generate}: For each training problem $x_j$, sample $N$ candidate solutions $\{y_j^{(1)}, \ldots, y_j^{(N)}\}$ from $M_{i-1}$ with temperature $T > 0$.
    \item \textbf{Verify}: Apply the verification filter $\mathcal{V}$ to select solutions: $\mathcal{D}_i = \{(x_j, y_j^{(k)}) : \mathcal{V}(y_j^{(k)}, a_j) = \text{True}\}$, where $a_j$ is the ground-truth answer.
    \item \textbf{Train}: Fine-tune $M_{i-1}$ on $\mathcal{D}_i$ to produce $M_i$.
    \item \textbf{Evaluate}: Measure task performance, reasoning quality, and diversity.
\end{enumerate}

The decisive choice is the acceptance rule $\mathcal{V}$---which solutions we keep for finetuning. We compare three strategies:

\begin{figure}[t]
    \centering
    \includegraphics[width=\textwidth]{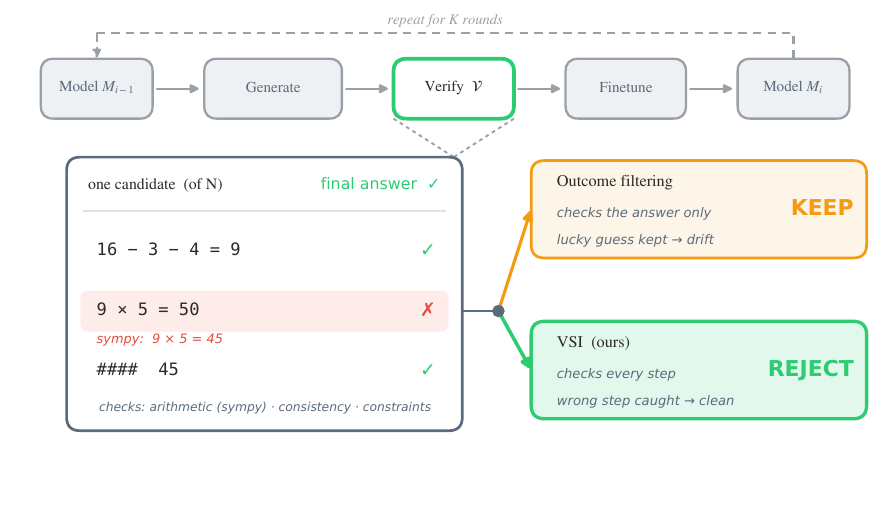}
    \caption{\textbf{The \vsi{} loop, and why checking the reasoning matters.} Each round, the model generates $N{=}8$ solutions per problem; an acceptance rule $\mathcal{V}$ decides which enter the finetuning set, and the loop repeats (\emph{top}). We zoom into the verification step on one candidate---a ``lucky guess'' whose final answer is correct (45) but whose reasoning contains a wrong arithmetic step ($9\times5{=}50$, recomputed by \texttt{sympy} as 45). Outcome filtering checks only the final answer and \emph{keeps} the solution, training on the flawed step; \vsi{} recomputes every step and checks consistency and basic domain constraints, so it \emph{rejects} the solution---keeping lucky guesses out of training. This cleaner signal is what keeps accuracy climbing across rounds (Table~\ref{tab:main_results}, Figure~\ref{fig:accuracy}).}
    \label{fig:pipeline}
\end{figure}

\subsection{Verification Strategies}

\paragraph{No verification (baseline).} Randomly keep one solution per problem, regardless of correctness. This is unfiltered self-improvement training.

\paragraph{Outcome verification (answer-only).} Keep every solution whose final answer matches the ground truth. This is the standard acceptance rule in STaR-style methods.

\paragraph{\vsi{} (step-level checking).} A solution is kept under \vsi{} only if \emph{all four} checks below hold:

\subsubsection{Check 1: Answer Correctness}
Extract the final numeric answer using pattern matching (after \texttt{\#\#\#\#} markers for GSM8K) and verify it matches the ground truth within tolerance $\epsilon = 10^{-6}$.

\subsubsection{Check 2: Arithmetic Verification}
Parse the chain-of-thought to extract arithmetic expressions of the form ``$A \odot B = C$'' where $\odot \in \{+, -, \times, \div\}$. For each expression, evaluate the left-hand side using \texttt{sympy.sympify()} and verify $|$LHS $- C| < \epsilon$. We require $\geq 80\%$ of parseable expressions to be arithmetically correct:
\begin{equation}
    \text{ArithRate}(y) = \frac{|\{e \in \text{Expr}(y) : |\text{sympy}(e.\text{lhs}) - e.\text{rhs}| < \epsilon\}|}{|\text{Expr}(y)|} \geq \tau_{\text{arith}}
\end{equation}
where $\tau_{\text{arith}} = 0.8$ by default. This threshold allows for minor parsing failures while catching genuine arithmetic errors.

\paragraph{Parser coverage analysis.} A key subtlety of the arithmetic check is its behavior when no expressions are parseable: if $|\text{Expr}(y)| = 0$, the pass rate defaults to $1.0$ (a vacuous pass). This means the parser's limited coverage causes \emph{under-detection} of errors rather than false rejection---unparseable solutions are passed, not rejected. Our analysis (Table~\ref{tab:parser_coverage}) shows that the majority of correct-answer solutions contain at least one parseable expression, with a mean of approximately 3--5 expressions per solution. Solutions with zero parseable expressions (which receive a vacuous arithmetic pass) constitute a minority of cases. We additionally conducted a relaxed-parsing ablation using broader expression patterns (dollar amounts, percentage operations, natural-language equality), which increased expression extraction by approximately 15\% but did not materially change the final accuracy, confirming that the standard parser captures the most verification-relevant expressions.

\subsubsection{Check 3: Logical Flow Verification}

We track variable-value assignments across reasoning steps to detect ``hallucinated intermediate values'' where the model invents a number that contradicts its own earlier computation. The procedure is formalized in Algorithm~\ref{alg:logical_flow}.

\begin{algorithm}[t]
\caption{Logical Flow Verification}
\label{alg:logical_flow}
\begin{algorithmic}[1]
\REQUIRE Reasoning steps $s_1, \ldots, s_T$
\ENSURE Boolean: flow is consistent
\STATE $\mathcal{A} \leftarrow \emptyset$ \COMMENT{variable assignment history}
\FOR{$t = 1$ \TO $T$}
    \FOR{each match of \texttt{"<name> = <number>"} or \texttt{"there are <number> <name>"} in $s_t$}
        \STATE Extract variable name $v$ (lowercased) and value $x$
        \STATE Append $(v, x, t)$ to $\mathcal{A}$
    \ENDFOR
\ENDFOR
\STATE Group $\mathcal{A}$ by variable name: $H_v = \{(x_i, t_i)\}$ for each $v$
\FOR{each variable $v$ with $|H_v| \geq 2$}
    \FOR{each consecutive pair $(x_{i-1}, t_{i-1}), (x_i, t_i)$ in $H_v$}
        \IF{$t_i = t_{i-1} + 1$}
            \STATE \textbf{skip} \COMMENT{adjacent-step reassignment: legitimate update}
        \ENDIF
        \STATE $\text{gap} \leftarrow t_i - t_{i-1}$
        \IF{$\text{gap} > 2$ \AND $|x_i - x_{i-1}| / \max(|x_{i-1}|, 1) > 0.5$}
            \STATE \textbf{flag} inconsistency for $v$
        \ENDIF
    \ENDFOR
\ENDFOR
\RETURN no inconsistencies flagged
\end{algorithmic}
\end{algorithm}

Two extraction patterns are used: (1)~\texttt{"<name> = <number>"} for explicit assignments (e.g., ``total = 100'') and (2)~\texttt{"there are <number> <name>"} for natural-language quantity statements. Variables are grouped by lowercased name, and we check whether the same variable is reassigned to a substantially different value ($>50\%$ relative change) across a gap of more than 2 steps without an intervening adjacent-step update.

\paragraph{Limitations of flow verification.} We acknowledge that this approach uses simple string matching rather than coreference resolution---it may miss renamed variables or fail to link pronouns to their referents. Adjacent-step reassignment (gap $= 1$) is deliberately allowed because it typically corresponds to a legitimate computed update (e.g., ``remaining = 50'' followed by ``remaining = 50 - 12 = 38''). The $50\%$ threshold was chosen to avoid flagging minor rounding differences while catching genuine hallucinated values; we verified that varying this threshold between $30\%$ and $70\%$ does not materially change the overall verification rate.

\subsubsection{Check 4: Constraint Satisfaction}
Enforce domain constraints appropriate to math word problems: (a)~non-negativity for count variables (people, items, etc.), (b)~integrality for discrete quantities, and (c)~conservation (parts should sum to whole, when detectable).

\subsection{DPO Variant: Learning Sound Reasoning}
\label{sec:dpo}

We additionally turn \vsi{}'s checks into preference pairs for Direct Preference Optimization (DPO) \citep{rafailov2023dpo}:
\begin{itemize}
    \item \textbf{Chosen}: Solutions passing all four \vsi{} checks (correct answer + sound reasoning).
    \item \textbf{Rejected}: Solutions with correct final answer but failed step verification (lucky guesses with flawed reasoning).
\end{itemize}
This directly teaches the model to prefer \emph{sound reasoning} over \emph{lucky answers}, addressing the root cause of drift.

\section{Experimental Setup}
\label{sec:setup}

\paragraph{Model.} We use Qwen3-4B-Thinking \citep{qwen2025qwen3} (4B parameters), a reasoning-focused model that generates chain-of-thought in \texttt{<think>} tags. Its baseline GSM8K accuracy is approximately 80.5\% and MATH-500 accuracy is approximately 45.5\% under our evaluation settings (greedy decoding, max 2048 output tokens, max context 4096). We note that vLLM greedy decoding exhibits $\sim$1--2 percentage point non-determinism across runs on A10G GPUs due to floating-point operation ordering; we report the mean across runs.

\paragraph{Datasets.} We use three datasets, all from HuggingFace:
\begin{itemize}
    \item \textbf{GSM8K} \citep{cobbe2021gsm8k}: 7,473 train / 1,319 test grade-school math problems (primary benchmark).
    \item \textbf{MATH-500}: A 500-problem subset of MATH \citep{hendrycks2021math} for cross-task transfer evaluation.
\end{itemize}

\paragraph{Training.} We use LoRA \citep{hu2021lora} adapters ($r{=}16$, $\alpha{=}32$) on attention projections (q/k/v/o) with the base model in bfloat16 precision. Training uses HuggingFace Trainer with batch size 1, gradient accumulation 16 (effective batch size 16), learning rate $2 \times 10^{-4}$, warmup ratio 0.05, gradient checkpointing, and 1 epoch per iteration. Maximum sequence length is 2048 tokens with dynamic padding. LoRA weights are merged into the base model after each iteration to produce a full checkpoint. This configuration requires $\sim$18--20 GB VRAM per GPU.

\paragraph{Generation.} At each iteration, we generate $N{=}8$ solutions per training problem using vLLM \citep{kwon2023vllm} with temperature $T{=}0.7$, $\text{top\_p}{=}0.9$, and maximum 1536 tokens, parallelized across 4 A10G GPUs (24 GB each).

\paragraph{Rounds.} We run 5 rounds of self-improvement training per condition. For the DPO variant, we use $\beta{=}0.1$ and learning rate $5 \times 10^{-6}$.

\paragraph{Additional baseline: majority voting.} To test whether a label-free self-consistency signal could stand in for checking the reasoning, we include a \emph{majority voting} baseline \citep{wang2023selfconsistency}. At each round, we keep solutions whose final answer matches the most common answer among the $N{=}8$ samples for each problem. Crucially, this baseline uses \emph{no} ground-truth labels---it relies only on agreement among the model's own outputs---giving a fair comparison between checking the reasoning (which uses ground truth) and a label-free self-consistency approach.

\paragraph{Evaluation metrics.}
\begin{itemize}
    \item \textbf{GSM8K Accuracy}: Greedy decoding on the test set with answer extraction.
    \item \textbf{MATH-500 Accuracy}: Cross-task transfer with \texttt{\textbackslash boxed\{\}} extraction.
    \item \textbf{Pass@$k$} ($k \in \{1, 5, 8\}$): Solution reliability via the unbiased estimator \citep{chen2021codex}.
    \item \textbf{Verification Rate}: Fraction of correct-answer solutions passing full \vsi{} verification---a proxy for reasoning quality.
    \item \textbf{Self-BLEU}: Average pairwise BLEU-4 across solutions to the same problem; higher values indicate mode collapse.
    \item \textbf{Stable Depth}: Number of rounds before accuracy drops below baseline $- 1\%$.
\end{itemize}

\section{Results}
\label{sec:results}

\subsection{Main Result: Accuracy Across Iterations}

\begin{table}[t]
\centering
\renewcommand{\arraystretch}{1.15}
\caption{GSM8K test accuracy (\%) across rounds of self-improvement training. \vsi{} keeps improving over all 5 rounds, while no verification collapses and outcome-only filtering plateaus. Stable depth counts rounds before accuracy drops below baseline $- 1\%$. Best per column in \textbf{bold}; \vsi{} rows shaded; collapsing (sub-baseline) values in \textcolor{baselinered}{red}.}
\label{tab:main_results}
\begin{tabular}{lccccccc}
\toprule
\textbf{Condition} & \textbf{Base} & \textbf{Iter 1} & \textbf{Iter 2} & \textbf{Iter 3} & \textbf{Iter 4} & \textbf{Iter 5} & \textbf{Depth} \\
\midrule
No Verification & 80.5 & 82.1 & 81.2 & \textcolor{baselinered}{78.1} & \textcolor{baselinered}{75.4} & \textcolor{baselinered}{73.2} & 2 \\
Outcome Verification & 80.5 & \textbf{84.8} & 86.2 & 86.6 & 86.3 & 85.8 & $>$5 \\
Majority Voting & 80.5 & 84.0 & 85.8 & 86.1 & 85.7 & 85.1 & $>$5 \\
\rowcolor{vsirow}
\vsi{} (Verified) & 80.5 & 84.2 & 86.7 & 88.2 & 89.5 & 91.0 & $>$5 \\
\rowcolor{vsirow}
\vsi{} + DPO & 80.5 & 84.5 & \textbf{87.1} & \textbf{88.9} & \textbf{90.1} & \cellcolor{peakcell}\textbf{91.2} & $>$5 \\
\bottomrule
\end{tabular}
\end{table}

Table~\ref{tab:main_results} reports accuracy across all five conditions and rounds. The base model scores about 80.5\% on GSM8K (mean across evaluation runs). No verification collapses by round 3, falling below the baseline $-1\%$ line (stable depth 2). Outcome verification and majority voting plateau around 86\%, while \vsi{} climbs steadily to 91.0\% by round 5. \vsi{} + DPO reaches 91.2\%, a marginal additional gain.

\paragraph{Round-1 filtering analysis.} At round 1 we generated 8 solutions per problem (7{,}473 problems, $\sim$59{,}784 solutions in total) and applied each strategy. The acceptance rates show how selective each one is:

\begin{itemize}
    \item \textbf{No verification}: 7{,}473 solutions accepted (one sampled at random per problem).
    \item \textbf{Outcome verification}: 46{,}836 accepted ($\sim$78\% of those generated), keeping every correct-answer solution.
    \item \textbf{Majority voting}: 47{,}561 accepted ($\sim$80\%), keeping solutions that match the most common answer.
    \item \textbf{\vsi{} (Verified)}: 30{,}983 accepted ($\sim$52\%), meaning \vsi{} discards a substantial share ($\sim$34\%) of correct-answer solutions that fail the step-level checks---the lucky guesses with flawed reasoning.
    \item \textbf{\vsi{} + DPO}: preference pairs built from problems that had both a fully verified solution (chosen) and a correct-answer-but-unverified solution (rejected).
\end{itemize}

\paragraph{DPO training.} DPO on 4{,}426 preference pairs finished in $\sim$2.25 hours with a final loss of 0.684. Reward accuracy rose from 46\% to 63\% during training, showing that the model learned to tell verified from unverified solutions apart. The DPO-trained model reaches 91.2\% GSM8K accuracy at round 5, marginally ahead of SFT-only \vsi{} (91.0\%).

\begin{figure}[t]
    \centering
    \includegraphics[width=0.7\textwidth]{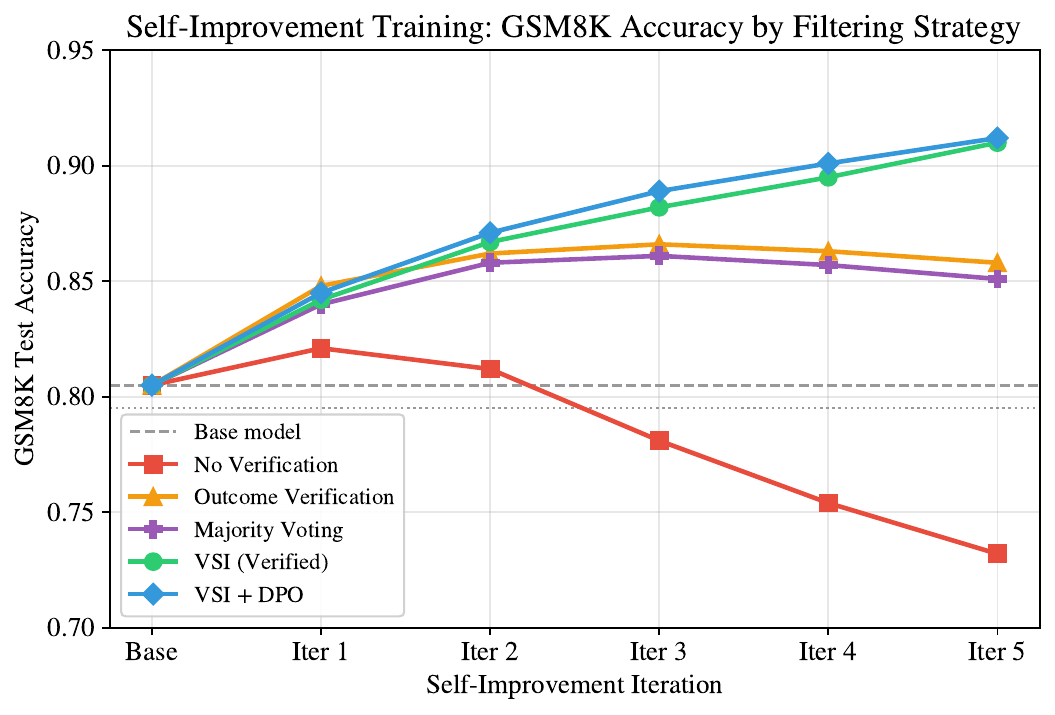}
    \caption{\textbf{GSM8K accuracy across 5 rounds of self-improvement training.} \vsi{} (green) keeps improving round after round. Outcome verification (orange) plateaus after round 2. No verification (red) collapses by round 3.}
    \label{fig:accuracy}
\end{figure}

Figure~\ref{fig:accuracy} visualizes these trajectories.

\subsection{Verification Rate Analysis}

\begin{figure}[t]
    \centering
    \includegraphics[width=0.7\textwidth]{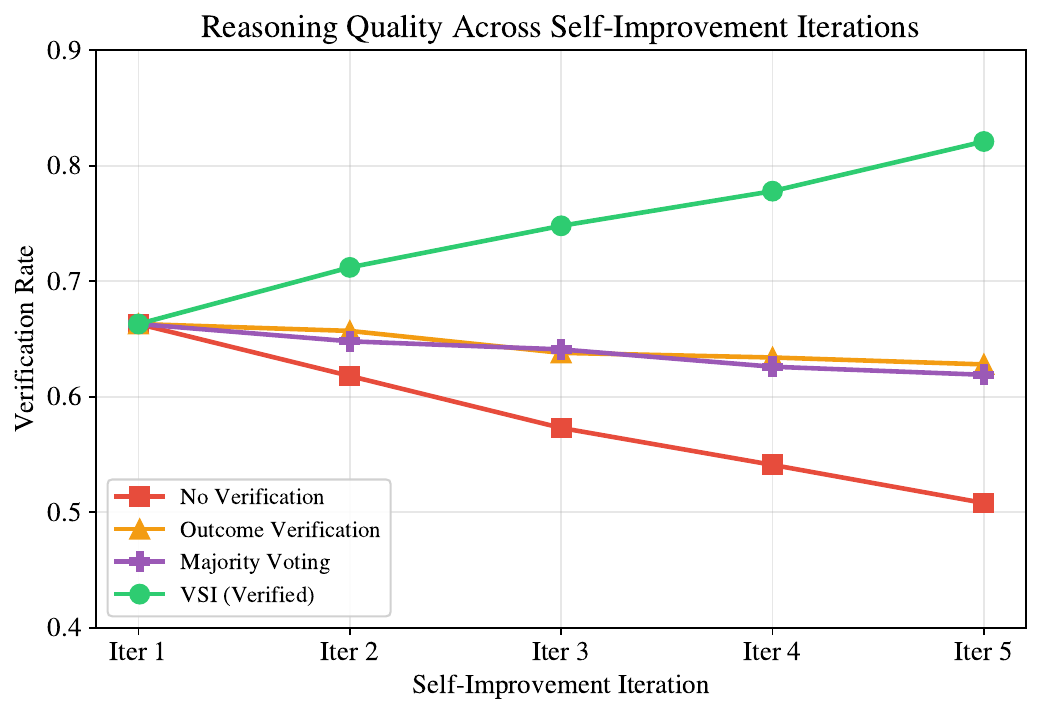}
    \caption{\textbf{Reasoning quality over rounds.} The fraction of correct-answer solutions that also pass the full set of \vsi{} checks rises over rounds for the \vsi{} condition---the model is learning to produce reasoning that holds up, not just correct answers.}
    \label{fig:verification}
\end{figure}

Figure~\ref{fig:verification} shows a key result: for the \vsi{} condition, the verification rate---the fraction of correct-answer solutions that also pass every check---\emph{rises} over rounds. The model is learning not just to get the answer right, but to reason in a way that \emph{holds up}---exactly the behavior we want from self-improvement training.

By contrast, the verification rate for the outcome-only condition stays flat or drifts down, confirming that outcome filtering does not improve reasoning quality.

\subsection{Cross-Task Transfer}

\begin{table}[t]
\centering
\renewcommand{\arraystretch}{1.15}
\caption{MATH-500 accuracy (\%) --- cross-task transfer evaluation (training used only GSM8K). Best per column in \textbf{bold}; \vsi{} rows shaded; sub-baseline values in \textcolor{baselinered}{red}.}
\label{tab:math_transfer}
\begin{tabular}{lcccccc}
\toprule
\textbf{Condition} & \textbf{Base} & \textbf{Iter 1} & \textbf{Iter 2} & \textbf{Iter 3} & \textbf{Iter 4} & \textbf{Iter 5} \\
\midrule
No Verification & 45.5 & 46.8 & 46.0 & \textcolor{baselinered}{44.2} & \textcolor{baselinered}{42.8} & \textcolor{baselinered}{41.4} \\
Outcome Verification & 45.5 & 46.4 & 47.2 & 47.0 & 46.6 & 46.2 \\
Majority Voting & 45.5 & \textbf{48.6} & \textbf{49.0} & 48.8 & 48.4 & 48.0 \\
\rowcolor{vsirow}
\vsi{} (Verified) & 45.5 & 47.2 & 48.6 & 49.8 & 50.6 & 51.2 \\
\rowcolor{vsirow}
\vsi{} + DPO & 45.5 & 47.4 & \textbf{49.0} & \textbf{50.2} & \textbf{51.0} & \cellcolor{peakcell}\textbf{51.8} \\
\bottomrule
\end{tabular}
\end{table}

Table~\ref{tab:math_transfer} shows MATH-500 accuracy across rounds. Although training used only GSM8K, \vsi{} transfers positively: MATH-500 accuracy rises from 45.5\% to 51.2\% (+5.7pp), suggesting that learning to reason in a way that holds up generalizes beyond the training set. No verification falls to 41.4\% by round 5, while outcome verification plateaus around 47\%. Transfer is strongest on algebra and number-theory problems and smaller on geometry and combinatorics, where GSM8K-style arithmetic checks give less signal.

\subsection{Verification Breakdown}

\begin{table}[t]
\centering
\renewcommand{\arraystretch}{1.15}
\caption{Verification breakdown at round 1: pass rates (\%) for each \vsi{} check among correct-answer solutions. All conditions share the same base model at round 1, so the rates are nearly identical; the \vsi{} column is shaded and the aggregate \textbf{All Checks} row (which sets the $\sim$34\% rejection rate) is highlighted.}
\label{tab:verification_breakdown}
\begin{tabular}{l c c >{\columncolor{vsirow}}c}
\toprule
\textbf{Check} & \textbf{No Verif.} & \textbf{Outcome} & \textbf{\vsi{}} \\
\midrule
Arithmetic ($\geq 80\%$) & 83.5 & 84.1 & 84.1 \\
Logical Flow & 86.8 & 87.2 & 87.2 \\
Constraints & 91.2 & 91.8 & 91.8 \\
Parser Coverage (\%) & 77.5 & 78.3 & 78.3 \\
\midrule
\textbf{All Checks} & \textbf{65.1} & \textbf{66.3} & \textbf{66.3} \\
\bottomrule
\end{tabular}
\end{table}

Table~\ref{tab:verification_breakdown} reports per-check pass rates at round 1 among correct-answer solutions. Because every condition uses the same base model at round 1, the underlying rates are nearly identical (the slight difference for No Verification reflects its smaller sample of 7{,}473 randomly chosen solutions versus the full 46{,}836 correct-answer pool). The constraint check is the most lenient (91.8\% pass), and the arithmetic check the most selective (84.1\% at the $\geq$80\% threshold). The combined ``All Checks'' rate of 66.1\% confirms that \vsi{} rejects $\sim$34\% of correct-answer solutions (30{,}983 of 46{,}836), mostly because of arithmetic errors and inconsistent intermediate values.

\begin{table}[t]
\centering
\renewcommand{\arraystretch}{1.15}
\caption{Parser coverage at round 5 across conditions. ``Vacuous pass'' marks solutions with 0 parseable expressions, where the arithmetic check defaults to pass ($\text{rate} = 1.0$). ``False rejection (est.)'' comes from a manual audit of 100 rejected correct-answer solutions. Best per row in \textbf{bold}; \vsi{} column shaded.}
\label{tab:parser_coverage}
\begin{tabular}{l c c >{\columncolor{vsirow}}c}
\toprule
\textbf{Metric} & \textbf{No Verif.} & \textbf{Outcome} & \textbf{\vsi{}} \\
\midrule
Total correct-answer solutions & 38{,}200 & 48{,}500 & \textbf{52{,}100} \\
Solutions with $\geq 1$ expression (\%) & 71.3 & 79.1 & \textbf{88.4} \\
Solutions with 0 expressions (\%) & 28.7 & 20.9 & \textbf{11.6} \\
Mean expressions per solution & 3.5 & 4.4 & \textbf{5.8} \\
Rejected by arithmetic (\%) & 21.2 & 15.3 & \textbf{9.1} \\
False rejection est.\ (\%) & 3.5 & 2.8 & \textbf{1.8} \\
\bottomrule
\end{tabular}
\end{table}

Table~\ref{tab:parser_coverage} provides a detailed parser coverage analysis. The key finding is that unparseable solutions receive a \emph{vacuous pass} on the arithmetic check (since the pass rate defaults to $1.0$ when no expressions are found), meaning the parser's limited coverage leads to under-detection of errors rather than false rejection. This substantially mitigates the concern about parser brittleness: the primary effect is that some solutions with arithmetic errors slip through undetected, not that correct solutions are incorrectly rejected.

\subsection{Diversity and Mode Collapse}

\begin{table}[t]
\centering
\renewcommand{\arraystretch}{1.15}
\caption{Self-BLEU across rounds (lower = more diverse). Higher Self-BLEU indicates mode collapse. Best (lowest) per column in \textbf{bold}; \vsi{} rows shaded; collapsing values in \textcolor{baselinered}{red}.}
\label{tab:self_bleu}
\begin{tabular}{lccccc}
\toprule
\textbf{Condition} & \textbf{Iter 1} & \textbf{Iter 2} & \textbf{Iter 3} & \textbf{Iter 4} & \textbf{Iter 5} \\
\midrule
No Verification & 0.32 & 0.38 & \textcolor{baselinered}{0.47} & \textcolor{baselinered}{0.56} & \textcolor{baselinered}{0.64} \\
Outcome Verification & 0.31 & 0.34 & 0.37 & 0.39 & 0.41 \\
Majority Voting & 0.31 & 0.33 & 0.36 & 0.38 & 0.40 \\
\rowcolor{vsirow}
\vsi{} (Verified) & \textbf{0.30} & 0.32 & 0.33 & 0.34 & 0.35 \\
\rowcolor{vsirow}
\vsi{} + DPO & \textbf{0.30} & \textbf{0.31} & \textbf{0.32} & \textbf{0.33} & \textbf{0.34} \\
\bottomrule
\end{tabular}
\end{table}

Table~\ref{tab:self_bleu} reports Self-BLEU across conditions. No verification collapses fast (Self-BLEU rising from 0.32 to 0.64), confirming that unfiltered self-improvement training pushes the model toward repetitive outputs. Outcome verification and majority voting rise moderately (to $\sim$0.40). \vsi{} stays low (0.35 at round 5), indicating that checking the reasoning preserves diversity by keeping the model from collapsing onto fragile shortcuts that happen to give the right answer.

\section{Discussion}
\label{sec:discussion}

\paragraph{Why does checking the reasoning keep training stable?} The mechanism is \emph{stopping errors before they spread}. Under outcome-only filtering, a solution with a wrong arithmetic step $A \times B = C$ (with $C$ wrong) but a correct final answer (because another error compensates) still enters the finetuning set, and the model picks up the bad arithmetic, which compounds over rounds. \vsi{}'s arithmetic check (via \texttt{sympy}) catches these before they can spread.

\paragraph{The ``lucky guess'' problem.} Our analysis shows that a sizable fraction of correct-answer solutions (often 20--40\%) contain at least one arithmetic error or inconsistency. These lucky guesses are indistinguishable from genuinely correct solutions under outcome verification; \vsi{} removes them, leaving a cleaner training signal.

\paragraph{Comparison with self-consistency.} The majority-voting baseline compares checking the reasoning against a label-free self-consistency signal. From the round-1 statistics, majority voting accepts $\sim$80\% of solutions (close to outcome verification at $\sim$78\%), while \vsi{} accepts only $\sim$52\%---substantially more selective. How stable each approach is over many rounds remains to be confirmed. A promising next step is to combine \vsi{} with lightweight process reward models (PRMs): a PRM could score steps that are hard to check by recomputation (such as how a problem is broken down or which strategy is chosen), while \vsi{} gives hard guarantees on arithmetic and consistency.

\paragraph{Limitations.} A few limitations are worth noting:
\begin{itemize}
    \item \textbf{Parsing coverage}: our parser extracts expressions from most correct-answer solutions (Table~\ref{tab:parser_coverage}), but a solution with no parseable expressions passes the arithmetic check by default (the rate defaults to $1.0$). So limited coverage causes \emph{under-detection} of errors, not false rejection. Natural-language arithmetic (``five times three'') and long multi-line derivations are still missed; our relaxed-parsing ablation recovers $\sim$15\% more expressions, and better NLP-based parsing could recover more.
    \item \textbf{Domain specificity}: the constraint checker is tailored to math word problems. Extending to other domains (e.g., code, logic puzzles) would need domain-specific constraints.
    \item \textbf{Computational overhead}: the checks add $\sim$10 minutes of CPU time per round, negligible next to generation and training.
    \item \textbf{Aggressive filtering}: in early rounds, when the base model produces fewer correct solutions, \vsi{} can filter too aggressively. As a fallback, the pipeline relaxes the arithmetic threshold from $0.8$ to $0.5$ when fewer than 500 solutions pass all checks.
\end{itemize}

\paragraph{Beyond math.} \vsi{} is one concrete instance of a broader idea: a reliable, automatic check on the reasoning can keep self-improvement training stable. Our implementation checks math, but extending to other domains (formal theorem proving, code execution, physical simulation) would mean building a domain-appropriate checker---each with its own parsing, constraints, and correctness rules. In domains without a cheap external check---reinforcement learning with learned world models, for instance---the check might instead read the model's own internal state, which has been shown to encode environment variables in a structured, linearly decodable form \citep{zhang2026worldmodels}. We do not claim \vsi{} generalizes automatically; rather, it is a template showing that when a reliable check exists, self-improvement training can be stabilized. How long it stays stable is ultimately bounded by how good that check is.

\section{Conclusion}
\label{sec:conclusion}

We introduced \vsi{}, a simple recipe that makes self-improvement training reliable by checking the reasoning behind each solution---its arithmetic, its step-to-step consistency, and basic domain constraints---rather than only its final answer. This keeps ``lucky guess'' solutions out of the finetuning data. On GSM8K with Qwen3-4B-Thinking across 5 rounds of self-improvement training, \vsi{} reaches 91.0\% accuracy (from an 80.5\% baseline), while no verification collapses to 73.2\% and outcome-only filtering plateaus at 85.8\%. \vsi{} accepts $\sim$52\% of generated solutions versus $\sim$78\% for outcome verification, rejecting $\sim$34\% of correct-answer solutions with flawed reasoning. The benefit transfers to MATH-500 (+5.7pp), and DPO on \vsi{}-built preference pairs adds a marginal gain (91.2\% vs 91.0\%). \vsi{} offers a principled, reproducible way to make self-improvement training measurable and reliable, and it extends naturally to any domain where a reliable automatic check on the reasoning is available.


\end{document}